\documentclass{article}

\usepackage{microtype}
\usepackage{graphicx}
\usepackage{subfigure}
\usepackage{booktabs} %

\usepackage{hyperref}

\usepackage{amsmath}
\usepackage{multirow}
\usepackage{amsthm}
\usepackage{bbm}

\usepackage{amssymb}

\usepackage[export]{adjustbox}

\theoremstyle{definition}
\newtheorem{definition}{Definition}
\newtheorem{example}{Example}

\usepackage[utf8]{inputenc} %
\usepackage[T1]{fontenc}    %
\usepackage{url}            %
\usepackage{booktabs}       %
\usepackage{amsfonts}       %
\usepackage{nicefrac}       %
\usepackage{microtype}      %
\usepackage{xcolor}         %
\usepackage{backref}

\usepackage{xspace}

\usepackage{bm}
\usepackage{float}
\usepackage{natbib}
\usepackage{graphicx}
\usepackage{algorithm}
\usepackage{algorithmic}
\usepackage{pdflscape}

\usepackage{amsmath,amsfonts,bm}

\def\eqref#1{equation~\ref{#1}}

\def\1{\bm{1}}

\DeclareMathAlphabet{\mathsfit}{\encodingdefault}{\sfdefault}{m}{sl}
\SetMathAlphabet{\mathsfit}{bold}{\encodingdefault}{\sfdefault}{bx}{n}

\newcommand{\R}{\mathbb{R}}

\mathchardef\ordinarycolon\mathcode`\:
\mathcode`\:=\string"8000
\begingroup \catcode`\:=\active
  \gdef:{\mathrel{\mathop\ordinarycolon}}
\endgroup

\newcommand{\name}{\textit{Syfer}\xspace}

\usepackage[accepted]{icml2021}

\icmltitlerunning{\name: Neural Obfuscation for Private Data Release}

\begin{document}

\twocolumn[
\icmltitle{ \name : Neural Obfuscation for Private Data Release}

\icmlsetsymbol{equal}{*}

\begin{icmlauthorlist}
\icmlauthor{Adam Yala}{equal,mit}
\icmlauthor{Victor Quach}{equal,mit}
\icmlauthor{Homa Esfahanizadeh}{mit}
\icmlauthor{Rafael~G.~L. D'Oliveira}{mit}
\icmlauthor{Ken R. Duffy}{may}
\icmlauthor{Muriel M\'{e}dard}{mit}
\icmlauthor{Tommi S. Jaakkola}{mit}
\icmlauthor{Regina Barzilay}{mit}
\end{icmlauthorlist}

\icmlaffiliation{mit}{Massachusetts Institute of Technology, USA}
\icmlaffiliation{may}{Maynooth University, Ireland}

\icmlcorrespondingauthor{Adam Yala}{adamyala@csail.mit.edu}
\icmlcorrespondingauthor{Victor Quach}{quach@csail.mit.edu}

\icmlkeywords{Machine Learning, ICML}

\vskip 0.3in
]

\printAffiliationsAndNotice{\icmlEqualContribution} %

\begin{abstract}

Balancing privacy and predictive utility remains a central challenge for machine learning in healthcare. In this paper, we develop \name, a neural obfuscation method to protect against re-identification attacks. \name composes trained layers with random neural networks to encode the original data (e.g. X-rays) while maintaining the ability to predict diagnoses from the encoded data. The randomness in the encoder acts as the private key for the data owner. We quantify privacy as the number of attacker guesses required to re-identify a single image (guesswork). We propose a contrastive learning algorithm to estimate guesswork. We show empirically that differentially private methods, such as DP-Image, obtain privacy at a significant loss of utility. In contrast, \name achieves strong privacy while preserving utility. For example, X-ray classifiers built with DP-image, \name, and original data achieve average AUCs of 0.53, 0.78, and 0.86, respectively.

\end{abstract}

\section{Introduction}

Data sharing is a key bottleneck for the development of equitable clinical AI algorithms.
Public medical datasets are constrained by privacy regulations \citep{HIPAA, GDPR}, that aim to prevent leakage of identifiable patient data.
We propose \name, an encoding scheme for private data release.
In this framework, data owners encode their data with a random neural network (acting as their private key) for public release. 
The objective is to enable untrusted third parties to develop classifiers for the target task, while preventing attackers from re-identifying raw samples.

An ideal encoding scheme would enable model development for arbitrary (i.e unknown) downstream tasks using standard machine learning tools. Moreover, this scheme would not require data owners to train their own models (e.g. a generative model).
Designing such an encoding scheme has remained a long-standing challenge for the community.
For example, differentially private methods pursue this goal by leveraging random noise to limit the sensitivity of the encoding to the input data. However, this often results into too large of a utility loss. 
In this work, we propose to learn a keyed encoding scheme, which exploits the asymmetry between the tasks of model development and sample re-identification, to achieve improved privacy-utility trade-offs.

The relevant notion of privacy, as defined by HIPAA, is \emph{de-identification}, i.e. preventing an attacker from identifying matching pairs between raw and encoded samples. We measure this risk using \emph{guesswork}, i.e. the number of guesses an attacker requires to match a single raw image to its corresponding encoded sample. 
We consider an extreme setting where the attacker has access to the raw images, the released encoded data and the randomized encoding scheme, and only needs to predict the matching between corresponding pairs (raw image, encoded image).
While the adversary can simulate the randomized encoding scheme, they do not have access to the data owner's private key.
Our evaluation setup acts as a worst-case scenario for data privacy, compared to a real-world setting where the attacker's knowledge of the raw images is imperfect.
To efficiently measure guesswork on real-world datasets, we leverage an model-based attacker trained to maximize the likelihood of re-identifying raw images across encodings. 

While an arbitrary distribution of random neural networks is insufficient to achieve strong privacy (i.e. high guesswork) on real-world datasets, we can learn to shape this distribution to obtain privacy on real data by composing random layers with trained \emph{obfuscator} layers. \name's obfuscator layers are optimized to maximize the re-identification loss of a model-based attacker on a public dataset while minimizing a reconstruction loss, maintaining the invertability of the whole encoding. To encode labels, we apply a random permutation to the label identities. 

We trained \name on a public X-ray dataset from NIH, and evaluated the privacy and utility of the scheme on heldout dataset (MIMIC-CXR) across multiple attacker architectures and prediction tasks. We found that \name obtained strong privacy, with an expected guesswork of 8411, i.e. when presented with a grid of 10,000 raw samples by 10,000 encoded samples, it takes an attacker an average of 8411 guesses to correctly guess a correct (raw~image, encoded~image) correspondence. Moreover, models built on \name encodings approached the accuracy of models built on raw images, obtaining an average AUC of 0.78 across diagnosis tasks compared to 0.84 by a non-private baseline with the same architecture, and 0.86 by the best raw-image baseline. In contrast, prior encoding schemes, like InstaHide~\cite{instahide} and Dauntless~\cite{DAUnTLeSS}, do not prevent re-identification, both achieving a guesswork of 1. While differential privacy schemes, such as DP-Image~\cite{liu2021dp}, can eventually meet our privacy standard with large enough noise, achieving a guesswork of 1379, this resulted in average AUC loss of 33 points relative to the raw-image baseline, i.e. an AUC of 0.53.

\section{Related Work}

\paragraph{Differentially Private Dataset Release}
Differential privacy \cite{dwork2014algorithmic} methods offer strong privacy guarantees by leveraging random noise to bound the maximum sensitivity of function outputs (e.g. dataset release algorithms) to changes in the underlying dataset. For instance, DP-Image \cite{liu2021dp} proposed to add laplacian noise to the latent space of an auto-encoder to produce differentially private instance encodings. Instead of directly releasing noisy data,  \cite{xie2018differentially, torkzadehmahani2019dp, jordon2018pate} propose to leverage generative adversarial networks (GANs), trained in a differentially private manner (e.g. DP-SGD \cite{abadi2016deep} or PATE  \cite{papernot2016semi}), to produce private synthetic data. However, differentially private GANs have been shown to significantly degrade image quality and result in large utility losses \cite{cheng2021can}. Instead of leveraging independent noise per sample to achieve privacy, \name obtains privacy through its keyed encoding scheme and thus enables improved privacy-utility trade-offs.

\paragraph{Cryptographic Techniques}
Cryptographic techniques, such as secure multiparty computation and fully homomorphic encryption~\cite{Yao86,GMW87,BGW88,DBLP:conf/stoc/ChaumCD88,Gentry09,BV14,cho2018secure} allow data owners to encrypt their data before providing them to third parties. These tools provide extremely strong privacy guarantees, making their encrypted data indistinguishable under chosen plaintext attacks (IND-CPA). However, building models with homomorphic encryption \cite{MohasselZ17,MiniONN,GAZELLE,crypto-2018-28796} requires leveraging specialized cryptographic primitives and induces a large computational overhead (ranging from 100x-10,000x \cite{lloret2021enabling}) compared to standard model inference. As a result, these tools are still too slow for training modern deep learning models. In contrast, \name considers a weaker threat model, where attackers cannot query the data owner's private-encoder (i.e no plaintext attacks) and our scheme specifically defends against raw data re-identification (the privacy notion of HIPAA). Moreover, \name encodings can be directly leveraged by standard deep learning techniques, improving their applicability.

\paragraph{Lightweight Encoding Schemes}
Our work extends prior research in lightweight encoding schemes for dataset release. Previous approaches \cite{Ko2020IeeeAccess,TanakaICCE2018,SirichotedumrongICIP2019} have proposed tools to carefully distort images to reduce their recognition rate by humans while preserving the accuracy of image classification models. However, these methods do not offer privacy against machine learning based re-identification attacks. \cite{wu2020privacy, xiao2020adversarial, wu2018towards, raval2019olympus} have proposed neural encoding schemes that aim to eliminate a particular private attribute (e.g. race) from the data while protecting the ability to predict other attributes (e.g. action) through adversarial training. 
These tools require labeled data for sensitive and preserved attributes, and cannot prevent general re-identification attacks while preserving the utility of unknown downstream tasks. Our work is most closely related to general purpose encoding schemes like InstaHide~\cite{instahide} and Dauntless~\cite{DAUnTLeSS,xiao2021art}. InstaHide encodes samples by randomly mixing images with MixUp~\cite{zhang2018mixup} followed by a random bitwise flip. Dauntless encodes samples with random neural networks and proved that the scheme offers strong information theoretic privacy if the input data distribution is Gaussian. However, we show that neither InstaHide nor Dauntless meet our privacy standard on our real-world image datasets. In contrast, \name leverages a composition of trained obfuscator layers and random neural networks to achieve privacy on real word datasets while preserving downstream predictive utility.

\paragraph{Evaluating Privacy with Guesswork }
Our study builds on prior work leveraging guesswork to characterize the privacy of systems \cite{Massey,Merhav,Arikan,Pfister,Beirami_tilting}. Guesswork quantifies the privacy of a system as the number of trials required for an adversary to guess private information, like a private key, when querying an oracle. In this framework, homomorphic encryption methods, which uniformly sample $b$-bit private keys, offer maximum privacy \cite{Calmon}, as the average number of guesses to identify the correct key is $2^{b-1}$. 
In the non-uniform guessing setting~\cite{Christiansen}, guesswork offers a worst-case notion of privacy  by capturing the situation where an attacker may only be confident on a single patient identity. Such privacy weaknesses are not measured by average case metrics, like Shannon entropy.

\section{Problem Statement}
\label{sec:problem_statement}

\begin{figure}[t]
    \centering
    \includegraphics[width=0.5\textwidth]{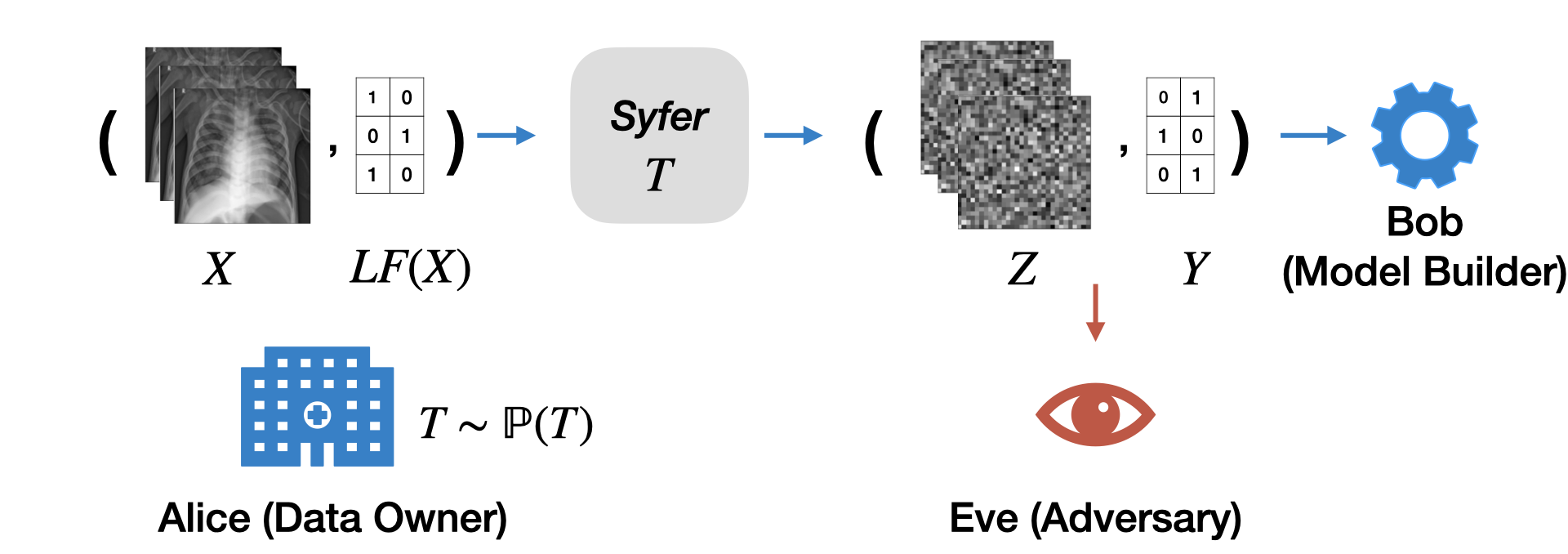}
    \caption{ Alice (data-owner) samples a transformation $T$ according to $\mathbb{P}(\bm{T})$  and leverages $T$ to encode her labeled data $(X, LF(X))$. She then publishes $(Z, Y) = T(X, LF(X))$ for model development by Bob, while preventing re-identification by Eve, the adversary.}
    \label{fig:overall}
\end{figure}

Our problem setting is illustrated in Figure \ref{fig:overall}.
A data owner (Alice) wishes to publish an encoded medical dataset to enable untrusted third parties (Bob) to develop machine learning models, while protecting patient privacy. We consider an adversary (Eve) with knowledge of Alice's image distribution, encoding scheme and encoded data. We measure the privacy of Alice's encoding scheme as the number of guesses it takes Eve to re-identify any single raw image from the encoded data.

\paragraph{Alice (the Data Owner).}
Alice uses a potentially randomized algorithm to transform each sample of her dataset and releases the encoded data along with their encoded labels.
Formally, let $X_A = \{x_1, ..., x_n\} \subset \R^d$ denote Alice's dataset, composed of $n$ samples, each of size $d$. She is interested in a $k$-class classification task defined by a labeling function $LF: \mathbb{R}^{d}  \rightarrow \{1, \dots , k\}$ (e.g. diagnosis labels annotated by a radiologist), and assembles the labeled dataset $\{(x, LF(x)), x \in X_A\}$.
To encode her data, Alice samples a transformation $T = (T^X, T^Y)$ according to a distribution $\mathbb{P}(\bm{T})$, where $T^X:\mathbb{R}^{d} \rightarrow \mathbb{R}^{d}$ is a sample encoding function and $T^Y:  \{1,...,k\} \rightarrow \{1,...,k\} $ is a label encoding function.
Alice releases the encoded dataset $(Z_A, Y_A) = T(X_A, LF(X_A)) = (T^X(X_A), T^Y(LF(X_A))$.
We use $\Gamma$ to denote the encoding scheme defined by $\mathbb{P}(\bm{T})$.

In the case of \name, Alice samples random neural network weights to construct a $T^X$ and samples a permutation to construct a $T^Y$.
We assume all actors have knowledge of the distribution $\mathbb{P}(\bm{T})$ (i.e. the neural network architecture and the pretrained weights of the obfuscator layers), but do not know Alice's specific choice of random $T$ (i.e. sampled weights or permutation).

\paragraph{Bob (the Model Builder).} Bob learns to classify the encoded data. He receives the encoded training dataset $(Z_A^{\text{train}},Y_A^{\text{train}})  = T(X_A^{\text{train}}, LF(X_A^{\text{train}}))$ and trains a model $C$ for the task of interest. Bob then shares the model $C$ with Alice who can then use the model on newly generated data and decode the predictions using the inverse of the label encoding, $\left(T^Y\right)^{-1}$. We note that Bob does not know $T$ or $\left(T^Y\right)^{-1}$. Bob's objective is to minimize the generalization error of his classifier on the testing set $(Z_A^{\text{test}}, Y_A^{\text{test}}) = T(X_A^{\text{test}}, LF(X_A^{\text{test}}))$.

\begin{definition}[utility]
    The \emph{utility score} of an encoding scheme $\Gamma$ for a labeling function $LF$ on a dataset $X_A$ is defined as the expected value of the generalization performance of Bob's classifier, i.e.
    \[
        \mathcal{U} = -\underset{T\sim \mathbb{P}(\bm{T})}{\mathbb{E}}\left[\underset{\tiny{x\in X_A^{\text{test}}}}{\mathbb{E}}\left[
        \mathcal{L}\left( \left(T^Y\right)^{-1} \circ C_T(z) , LF(x)\right)
        \right]\right]
    \]
\end{definition}
    
where $\mathcal{L}$ is a loss function, $z=T^X(x)$, $C_T$ is a classifier trained for a specific transformation $T$, and $\left(T^Y\right)^{-1}$ is the inverse of the label encoding.

\paragraph{Eve (the Adversary).}
Eve's objective is to re-identify Alice's data. In a realistic setting, Eve would use an encoded image from Alice's released dataset to reconstruct the pixels of the corresponding raw image.
To simplify analysis, we consider an easier task for Eve where she only needs to select the correct raw image from a list $X_E$ of candidate raw images, akin to identifying a suspect from a police lineup.
In the rest of the paper, we use $X_E = X_A$.
Knowing the dataset $X_A$ and the encoding scheme $\Gamma$,
Eve's task is to leverage the released dataset $(Z_A, Y_A) = T(X_A, LF(X_A))$ to re-identify samples from $Z_A$. Specifically, Eve's objective is to identify a \emph{single} correct match within $M_T = \{ (x, z) \in X_A \times Z_A   \text{ s.t. } z = T^X(x) \}$. %

\begin{definition}[privacy]
    We quantify the \emph{privacy} of an encoding scheme by computing the \emph{guesswork} of its transformations.
    Intuitively, a computationally unbounded Eve ranks pairs of $(x,z)$ as most likely to least likely. Guesswork is defined as the rank of Eve's first correct guess.
    Given the dataset $X_A$, the encoding scheme $\Gamma$, i.e. $\mathbb{P}(\bm{T})$, the released dataset $Z_A$ (of size $n$) along with released labels $Y_A$,  Eve derives for any $(x,z) \in X_A \times Z_A$ the probability $\mathbb{P}\left( (x,z) \in M_T | Z_A, Y_A, X_A, \Gamma \right)$. Eve then submits an ordered list of $n^2$ correspondence guesses $( u_1, u_2, \dots u_{n^2})$, where  $u_i \in  X_A \times Z_A$, by greedily
        ordering\footnote{Whenever ties occur, we compute $\mathcal{G}(T)$ as an average over the permutations of the list that keep it ordered.
    }
    her guesses from most likely to least likely. The guesswork of a transformation $T$ is defined as the index of the first correct guess in the ordered list, i.e.
    \[
    \mathcal{G}(T) = \min_k \{k \text{ s.t. } u_k \in M_T\}
    \text{.}\]
    Finally, to compare the privacy of encoding schemes, we compare the distributions of $\mathcal{G}(T)$ as $T$ is drawn from different distributions $\mathbb{P}(\bm{T})$.

\end{definition}

\begin{example}[guesswork calculation]
    Let $X_E = X_A = \{x_1, x_2\}$ and disregard labels for now. Consider two distinct encoders $T_1$ and $T_2$ that transform symbols in $X_A$ to $\{z_1,z_2\}$ as follows:

    $
   T_1: \left\{
        \begin{array}{ll}
            x_1 & \mapsto z_1 \\
            x_2 & \mapsto z_2
        \end{array}
    \right.
    \quad
    T_2: \left\{
        \begin{array}{ll}
            x_1 & \mapsto z_2 \\
            x_2 & \mapsto z_1
        \end{array}
    \right.
    $

    We evaluate an encoding scheme $\Gamma$, defined by the distribution used to sample $T$:
    $\quad
    \mathbb{P}(T_1) = 2/3,
    \quad \mathbb{P}(T_2) = 1/3$.

   Regardless of Alice's choice of $T$, Eve observes $Z_A = \{z_1, z_2\}$. Given her knowledge of $\mathbb{P}(\bm{T})$, she elects to rank  $(x_1, z_1)$ and $(x_2,z_2)$ before $(x_1,z_2)$ and $(x_2,z_1)$, which gives guessworks $\mathcal{G}(T_1) = 1$ and $\mathcal{G}(T_2) = 3$. In expectation, the guesswork of $\Gamma$ is 5/3.

   In general, we show in Appendix~\ref{sec:guesswork} that the guesswork of private schemes linearly scales with the size of $X_A$ and a uniform $\mathbb{P}(x,z)$ leads to a guesswork of $\frac{n^2+1}{n+1}$.
\end{example}

\section{Method}

We propose \name, an encoding scheme which uses a combination of learned \emph{obfuscator} layers and random neural network layers to encode raw data. 
\name is trained to maximize the re-identification loss of an attacker while minimizing a reconstruction loss, which acts as a regularizer to preserve predictive utility for downstream tasks.  
To estimate the privacy of an encoding scheme on a given dataset, we use a model-based attacker trained to maximize the likelihood of re-identifying raw data. 
To encode the labels $LF(X)$, \name randomly chooses a permutation of label identities $\{1,...,k\}$.

\subsection{Privacy Estimation via Contrastive Learning}
\label{priv_estim}

\begin{figure}[t]
    \centering
    \includegraphics[width=0.5\textwidth]{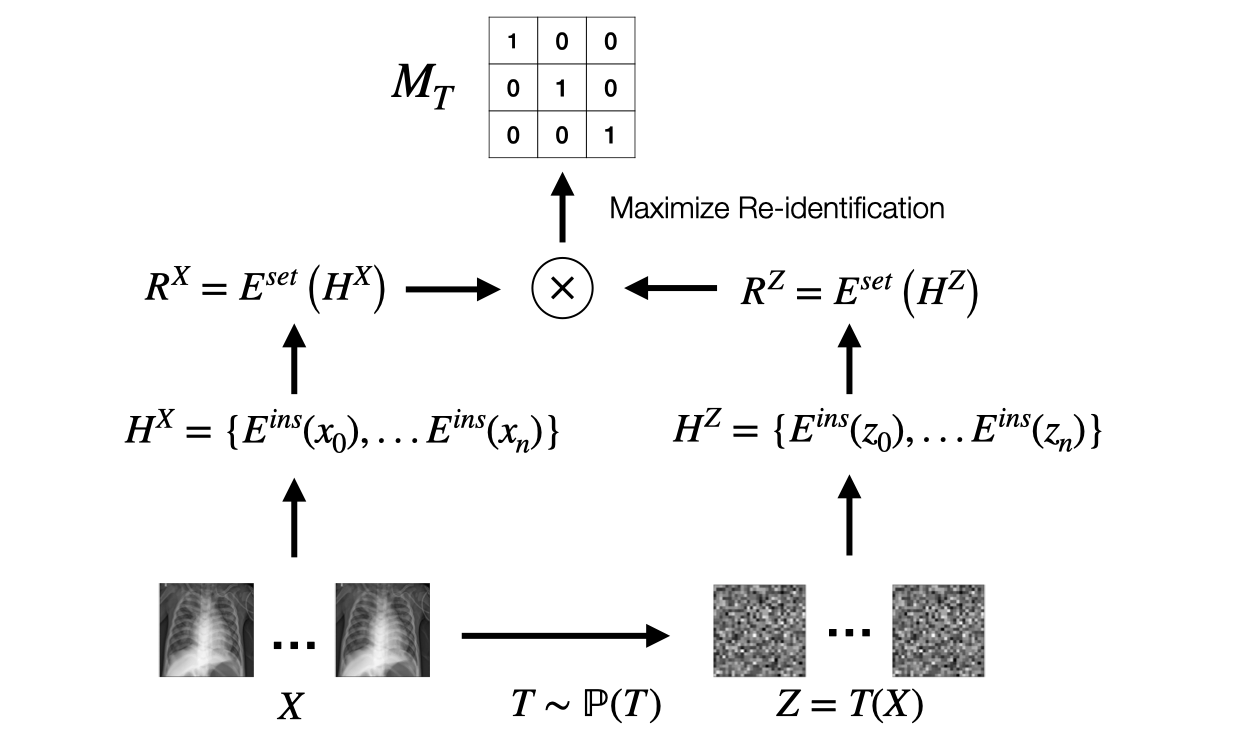}
    \caption{Architecture of the model-based attacker. Given pairs of raw samples $(X, LF(X))$ and encoded samples $(Z,Y)$, the attacker learns to recover matching pairs $(x,z) \in M_T$.
    In this figure, we omit label information for clarity.}
    \label{fig:contrastive}
\end{figure}

Before introducing \name, we adopt Eve's perspective and describe how to evaluate the privacy of encoding schemes. 
The attacker is given the candidate list $X_E = X_A$, and a fixed encoding scheme $\Gamma$, i.e. a fixed distribution $\mathbb{P}(\bm{T})$. We propose an efficient contrastive algorithm to estimate $\mathbb{P}\left( (x,z) \in M_T | Z_A, Y_A, \Gamma, X_A \right)$.
When the context allows it, we omit the conditional terms and use $\mathbb{P}( (x,z) \in M_T)$.

As shown in Figure~\ref{fig:contrastive}, the attacker's model $E$ is composed of an instance-level encoder $E^{\text{ins}}$, with parameters $\varphi^{\text{ins}}$, acting on individual images and their labels and a set-level encoder $E^{\text{set}}$, with parameters $\varphi^{\text{set}}$, taking a set of instance representations as input. 

In each iteration, we sample a batch $X = (x_1, \dots, x_b)$ of datapoints from $X_E = X_A$ and a transformation $T=(T^X, T^Y)$ according to the fixed distribution $\mathbb{P}(\bm{T})$.
Let $Z = T^X(X) = (z_1, \dots, z_b)$ denote the transformed batch and $Y = T^Y(LF(X)) = (y_1, \dots, y_b)$ the encoded labels.
The hidden representations of the raw data are computed as a two-step process:

\begin{enumerate}
    \item using $E^{\text{ins}}$, we compute $H^X = (h^X_1,\dots, h^X_b)$ where each $h^X_i = E^{\text{ins}}\left(x_i, LF(x_i)\right)$  ;
    \item using $E^{\text{set}}$, we compute $R^X = (r^X_1,..., r^X_b)$ where each $r^X_i = E^{\text{set}}\left(h^X_i, H^X\right)$.
\end{enumerate}

Similarly, for the encoded data, we form $H^Z = (h^Z_1,\dots, h^Z_b)$ where $h^Z_i = E^{\text{ins}}\left(z_i, y_i\right)$ and $R^Z = (r^Z_1,..., r^Z_b)$ where each $r^Z_i = E^{\text{set}}\left(h^Z_i, H^Z\right)$.
 
Following prior work on contrastive estimation \citep{chen2020simple}, we use the cosine distance between hidden representations to measure similarity:
\[
\text{sim}(r^X_i, r^Z_j) =\frac{\left(r^X_i\right)^{\top} r^Z_j }{ \Vert r^X_i \Vert \Vert  r^Z_j \Vert}
\text{ .}\]
Then, we estimate the quantity $\mathbb{P}( (x_i,z_j) \in M_T)$ as proportional to $\hat{p}(x_i, z_j)$:

\[
\hat{p}(x_i, z_j) =  \frac{\exp(\text{sim}(r^X_i, r^Z_j))}{\sum_{k,l}^b \exp(\text{sim}(r^X_k, r^Z_l))}
\text{ .}\]

The weights $\varphi^{\text{ins}}$ and $\varphi^{\text{set}}$ of the attacker's model $E$ are trained to minimize the negative log-likelihood of re-identification across unknown $T$: 
\[
\mathcal{L}_{\text{reid}} = -\sum_{(x,z) \in M_T} \log\left( \hat{p}(x, z) \right) 
\text{ .}\]

\subsection{\name } %
\paragraph{Architecture}

\newcommand{\thetakey}{\theta_{{key}}}
\newcommand{\thetakeybatch}{\thetakey^\text{batch}}
\newcommand{\thetaobf}{\theta_{\text{\name}}}

\begin{figure}[t]
    \centering
    \includegraphics[width=0.5\textwidth]{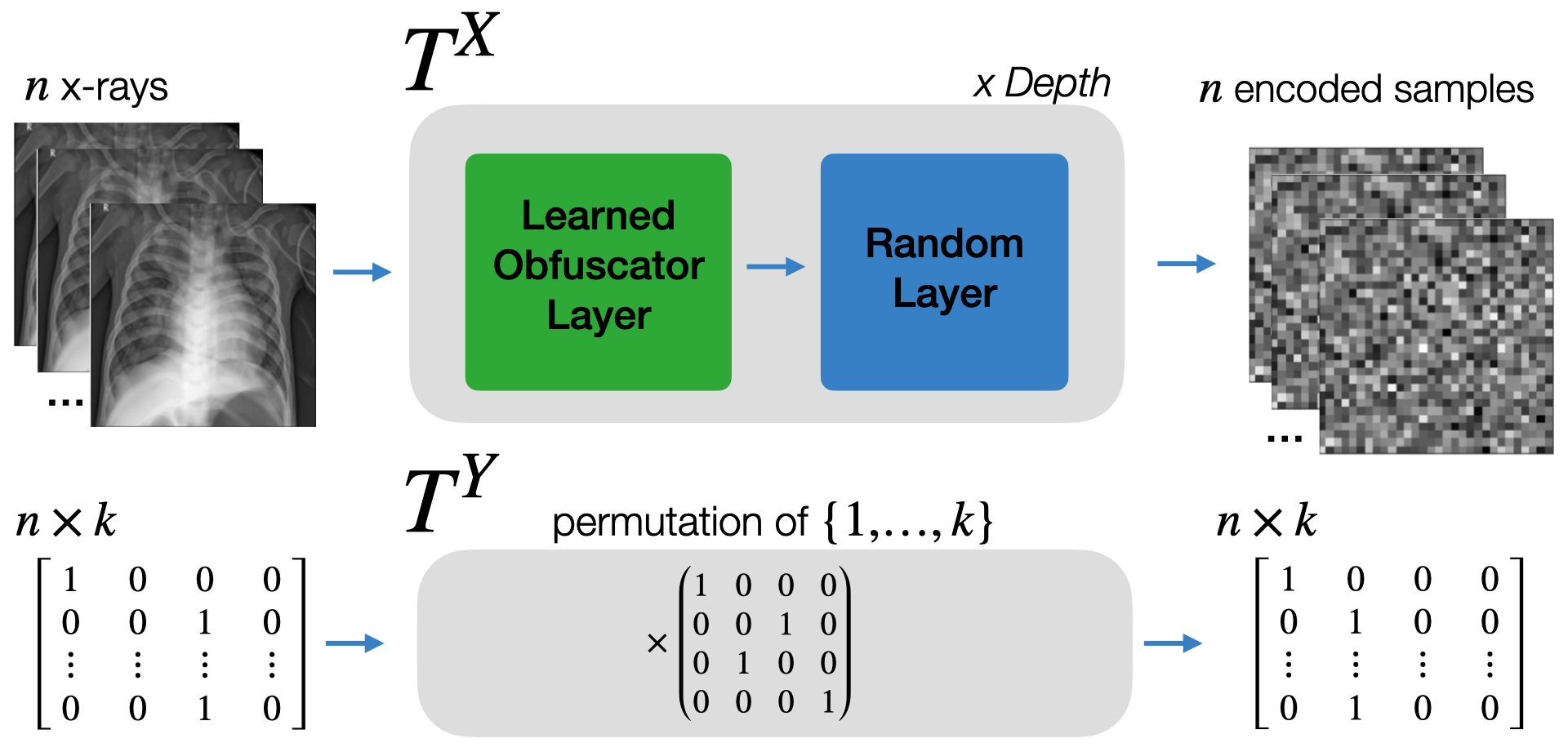}
    \caption{Proposed encoding scheme: \name uses repeating blocks of learned obfuscator layers and random neural network layers as $T^X$ and samples a random permutation of $\{1,...,k\}$ as $T^Y$.}
    \label{fig:syfer-arch}
\end{figure}

As illustrated in  Figure \ref{fig:syfer-arch},  we propose a new encoding scheme by learning to shape the distribution $\mathbb{P}(\bm{T})$. Specifically, we parametrize a transformation $T^X$ using a neural network that we decompose into blocks of learned \emph{obfuscator} layers (weights $\thetaobf$), and random layers (weights $\thetakey$).
The \emph{obfuscator} layers are trained to leverage the randomness of the subsequent random layers and learn a distribution $\mathbb{P}(\bm{T})$ that achieves privacy.
In this framework, Alice constructs $T^X$ by randomly sampling the weights $\thetakey$ and composing them with pre-trained obfuscator weights $\thetaobf$ to encode the raw data $X$. 
Alice chooses the label encoding $T^Y$ by randomly sampling a permutation of the label identities $\{1, \dots, k\}$, which is applied to $LF(X)$.
We note that our $T^Y$ assumes that Alice's dataset is class-balanced\footnote{If Alice's data is not class-balanced, she can  down-sample her dataset to a class-balanced subset before release.}. %

Alice's random choices of $\thetakey$ and $T^Y$ act as her private key, and she can publish the encoded data with diagnosis labels for model development while being protected from re-identification attacks. Given Bob's trained classifier to infer $T^Y(LF(x))$ from $T^X(x)$, Alice then uses $\left(T^Y\right)^{-1}$ to decode the predictions.

\paragraph{Training}

\begin{algorithm}[ht]
\caption{\emph{Syfer} training}
\label{alg:train}
\fontsize{9.5}{11.4}\selectfont
\begin{algorithmic}[1]
\STATE Initialize obfuscator parameters $\thetaobf$
\STATE Initialize attacker $E$ with parameters $\varphi = (\varphi^{\text{ins}}, \varphi^{\text{set}})$
\STATE Initialize decoders $D_1, \dots D_s$ with parameters $\beta_1, \dots, \beta_s$
\STATE For each decoder, sample random layer weights\\ $\thetakey^1, \dots \thetakey^s$ (fixed throughout training)
\STATE Set flag \textit{optimize\_estimators} $\gets$ true
\REPEAT
 	\STATE Sample a batch of datapoints $X$ from $X^{\text{public}}$
 	\vspace{2pt}
    \STATE \colorbox{lightgray}{$\triangleright$ Step 1: Compute re-identification loss}
    \STATE Sample a set of random layer weights $\thetakeybatch$ 
    \STATE Using obfuscator parameters $\thetaobf$ and key $\thetakeybatch$:
    \STATE $T^{\text{batch}} \gets f(\thetaobf, \thetakeybatch)$
    \STATE 
    $\left(Z^{\text{batch}}, Y^{\text{batch}}\right) \gets T^{\text{batch}}(X, LF(X)) $
    \STATE
    $R^Z \gets E_{\varphi}\left(Z^{\text{batch}}, Y^{\text{batch}}\right)$ 
    \STATE $R^X \gets E_{\varphi}(X,LF(X))$ 
    \STATE $\mathcal{L}_{\text{reid}} \gets \text{contrastive\_loss} \left(R^X, R^Z\right)$
    \STATE \colorbox{lightgray}{$\triangleright$ Step 2: Compute reconstruction loss}
 	\STATE  $\mathcal{L}_{\text{rec}} \gets 0$
    \FOR{$i \in \{1,\dots s\}$}
        \STATE Using  obfuscator parameters $\thetaobf$ and fixed key  $\thetakey^i$: \\
        \STATE $T^{i} \gets f(\thetaobf, \thetakey^i)$
        \STATE  $\left(Z^{i},Y^{i}\right) \gets T^{i}(X, LF(X)) $
     	\STATE  $\mathcal{L}_{\text{rec}} \gets \mathcal{L}_{\text{rec}} + \text{MSE}\left ( D_i\left(Z^{i}\right), X\right)$
 	\ENDFOR
    \STATE \colorbox{lightgray}{$\triangleright$ Step 3: Alternatively update parameters}
 	\IF {\textit{\textit{optimize\_estimators}}}
     	\STATE $\varphi \gets  \varphi - \nabla_{\varphi} \mathcal{L}_{\text{reid}}$
      	\STATE $\beta_i \gets \beta_i - \nabla_{\beta_i} \mathcal{L}_{\text{rec}}$ \quad \COMMENT{for $i \in \{1,\dots s\}$}
     	\STATE \textit{\textit{optimize\_estimators}} $\gets$ false
 	\ELSE
 	    \STATE $\thetaobf \gets \thetaobf -\nabla_{\thetaobf} (\lambda_{\text{rec}} \cdot \mathcal{L}_{\text{rec}} - \lambda_{\text{reid}} \cdot \mathcal{L}_{\text{reid}})$
 	    \STATE \textit{\textit{optimize\_estimators}} $\gets$ true
 	\ENDIF
\UNTIL{convergence}
\end{algorithmic}
\end{algorithm}

Data owners may not have the computational capacity to train their own obfuscator layers, so we train \name without direct knowledge of $X_A$ or $LF$. Instead, we rely on a public dataset  $X^{\text{public}}$ and use the null labeling function $LF(x)=0$.
To be successful, \name needs to generalize to held-out datasets, prediction tasks and attackers.

As shown in Algorithm~\ref{alg:train}, we train \name's obfuscator layers (parameters $\thetaobf$) to maximize the loss of an attacker $E$ (parameters  $\varphi = (\varphi^{\text{ins}}, \varphi^{\text{set}})$) and to minimize the reconstruction loss of an ensemble of decoders $D_1, \dots, D_s$ (parameters $\beta_1, \dots \beta_s$). 

At each step of training, we sample a transformation $T^{\text{batch}}$ by choosing a new $\thetakeybatch$ to combine with the current $\thetaobf$ (Alg.~\ref{alg:train}, L.9-11). Using the current attacker weights $\varphi$, we then compute the re-identification loss (Alg.~\ref{alg:train}, L.13-15) as: 
\[
\mathcal{L}_{\text{reid}} = -\sum_{(x,z) \in M_T } \log\left( \hat{p}(x, z) \right) 
\]

Next, we estimate the overall invertability of the encoding scheme by measuring the reconstruction loss of an ensemble of decoders $D_1, \dots, D_s$.
For each each decoder $D_i$, we randomly sample a private key $\thetakey^i$, which is fixed throughout the training algorithm.
Each decoder $D_i$ is trained to reconstruct $X$ from $Z = T^i(X)$ where $T^i$ is constructed by composing  the current $\thetaobf$ with $\thetakey^i$. We update $\beta_i$ to minimize the reconstruction loss (Alg.~\ref{alg:train}, L.17-23):

$$
    \mathcal{L}_{\text{rec}} =  \sum_{i=1}^s\left( \mathbb{E}_X[ ||x  - D_i \circ T^i(x) ||^2]  \right)
$$

We train our attacker and decoders in alternating fashion with \name's obfuscator parameters. 
On even steps, \name's weights $\thetaobf$ are updated to minimize the loss:
$$\mathcal{L}_{\name} = \lambda_{\text{rec}} \cdot \mathcal{L}_{\text{rec}} - \lambda_{\text{reid}} \cdot \mathcal{L}_{\text{reid}} $$
On odd steps, the attacker and decoders are updated to minimize $\mathcal{L}_{\text{reid}}$ and $\mathcal{L}_{\text{rec}}$ respectively (Alg.~\ref{alg:train}, L. 25-32).

In this optimization, the tasks of our attacker and decoders are asymmetric: the attacker is trained to generalize across transformations $T$ (i.e. $\thetakey$), while the decoders only need to generalize to unseen images, for a fixed key $\thetakey$.

\section{Experiments}

\textbf{Datasets} \quad For all experiments, we utilized two benchmark datasets of chest X-rays, NIH \cite{wang2017chestx} and MIMIC-CXR \citep{johnson2019mimic}, from the National Institutes of Health Clinical Center and Beth Israel Deaconess Medical Center respectively. Both datasets were randomly split into 60,20,20 for training, development and testing, and all images were downsampled to 64x64 pixels. We leveraged the NIH dataset to train all private encoding schemes (i.e. \name and baselines), and we evaluated the privacy and utility of all encoding schemes on the MIMIC-CXR dataset, for the binary classification tasks ($k=2$) of predicting Edema, Consolidation, Cardiomegaly, and Atelectasis. This reflects the intended use of the tool, where a hospital leverages a pretrained \name for their heldout datasets.

For privacy and utility experiments considering a specific diagnosis task, we used a filtered version of the MIMIC-CXR data with balanced labels and explicit negatives. Specifically, for each diagnosis tasks, we followed common practice~\cite{irvin2019chexpert} and excluded exams with an uncertain disease label, i.e., the clinical diagnosis did not explicitly rule out or confirm the disease. Then, we selected one random negative control case for each positive case in order to create a balanced dataset.
Our dataset statistics are shown in Appendix \ref{sec:dataset_stats}.

\textbf{\name Implementation Details}\quad  As shown in Figure \ref{fig:syfer-arch}, \name consists of repeated blocks of trained obfuscator layers and random neural network layers. Following prior work in vision transformers~\citep{zhou2021deepvit}, \name operates at the level of patches of images. We used a patch size of 16x16 pixels and 5 \name blocks for all experiments. We implemented our trained obfuscator layers as Simple Attention Units (SAU), a gated multi-head self attention module. We implemented our random neural networks as linear layers, followed by a SeLU nonlinearity and layer normalization. All random linear layers weights were sampled from a unit Gaussian, and we used separate random networks per patch. Our full \name architecture has 12.9M parameters, of which 6.6M are learned obfuscator parameters and 6.3M are random neural layer parameters.
The SAU module is detailed in Appendix~\ref{sec:sau}.

We trained \name for 50,000 steps on the NIH training set to maximize the re-identification loss and minimize the reconstruction loss, with $\lambda_{\text{reid}}=2$, $\lambda_{\text{rec}}=20$. We trained our adversary and decoder for one step for each step of obfuscator training. We implemented the instance encoder $E^{\text{ins}}$ and set encoder $E^{\text{set}}$ of our adversary model as a depth 3 and depth 1 SAU respectively. We utilized separate $E^{\text{ins}}$ networks to encode the raw data $(X, LF(X))$ and encoded data $(Z, Y)$. We use a single decoder\footnote{Using an ensemble of $s=5$ decoders did not significantly improve downstream utility.} $D_1$ (i.e. $s=1$) and implement it as a depth 3 SAU. We used a batch size of 128, the Adam optimizer and a learning rate of $0.001$ for training \name and our estimators. The training of \name is fully reproducible in our code release.

\paragraph{Privacy Estimators}
To evaluate the ability of \name to defend against re-identification attacks, we trained attackers to re-identify raw images from \name encodings on the MIMIC-CXR dataset.
Since we cannot bound the prior knowledge the attacker may have over $X_A$, we
consider the extreme case and train our attackers on their evaluation set, i.e. we only use MIMIC-CXR's training set for privacy evaluation.
As a result, the attacker does not have to generalize to held-out images, but only to held-out private encoders $T$.

As described in Section \ref{priv_estim}, the attacker is trained to re-identify raw images from encoded images across new unobserved private keys using an image encoder $E^{\text{ins}}$ and a set encoder $E^{\text{set}}$. This attacker estimates $\mathbb{P}( (x,z) \in M)$ for an encoding scheme $\Gamma$ on a dataset $X$. Across our experiments, we implemented $E^{\text{ins}}$ as either a ResNet-18~\cite{he2016deep}, a ViT ~\cite{zhou2021deepvit}, or a SAU. We implemented $E^{\text{set}}$ as a depth 1 SAU. All attackers were trained for 500 epochs.

We computed the guesswork of each attacker by sorting the scores $\hat{p}(x,z)$ and identifying the index of the first correct correspondence. To measure the attackers average performance, we also evaluated the ROC AUC of the attacker attempting to predict an $(x,z)$ matching as a binary classification task.  A higher guesswork and lower re-identification AUC (ReID AUC) reflect a more private encoding scheme.

\begin{table}[!t]
\centering

\begin{tabular}{lll}
\hline
\textit{Encoding} & Guesswork&  ReId AUC\\
\hline
Dauntless          & $1.0 $  $(1, 1)       $ & $1.00$ $(1.00, 1.00)$ \\
InstaHide          & $1.0 $  $(1, 1)       $ & $1.00$ $(1.00, 1.00)$ \\
DP-S, $b=10$       & $1.2 $  $(1, 2)       $ & $0.98 $ $(0.98, 0.98)  $\\
DP-S, $b=20$       & $7.2 $  $(1, 31)      $ & $0.86 $ $(0.85, 0.86)  $\\
DP-S, $b=30$       & $68  $  $(1, 205)     $ & $0.70 $ $(0.70, 0.70)  $\\
DP-I, $b=1$        & $5.0 $  $(1, 17)      $ & $0.89 $ $(0.88, 0.89)  $\\
DP-I, $b=3$        & $77  $  $(3, 276)     $ & $0.73 $ $(0.73, 0.73)  $\\
DP-I, $b=5$        & $1379$  $(49, 4135)   $ & $0.59 $ $(0.59, 0.60)  $\\
\name-Random       & $1.7 $  $(1, 4)       $ & $0.99 $ $(0.99, 0.99) $\\
\name ($T^X$ only) & $8476 $  $(1971, 20225)$ & $0.50 $ $(0.49, 0.52) $\\
\hline

\end{tabular}
\caption{Privacy evaluation of different encoding schemes against an SAU based attacker on the unlabeled MIMIC-CXR dataset. 
For \name, only $T^X$ is used. 
DP-S and DP-I stand for DP-Simple and DP-Image respectively.
The scale parameter $b$ characterizes the laplacian noise.
Metrics are averages over 100 trials using 10,000 samples each, followed by 95\% confidence intervals (CI). 
}%
\label{tab:gen-privacy}
\end{table}

\begin{table}
\centering

\begin{tabular}{lll}
\hline
Attacker & Guesswork &  ReId AUC\\ 
\hline
SAU       & $8476 $ $(1971, 20225)$ & $0.50$ $(0.49, 0.52)$ \\
ViT       & $8411 $ $(5219, 12033)$ & $0.50$ $(0.49, 0.51)$ \\
Resnet-18 & $10070$ $(9871, 10300)$ & $0.50$ $(0.47, 0.53)$  	 \\
\hline
\end{tabular}
\caption{Privacy evaluation of \name across different attacker architectures on the unlabeled MIMIC-CXR dataset. 
Metrics are averages over 100 trials using 10,000 samples each, followed by 95\% CI. 
}%
\label{tab:cross-attacker-privacy}
\end{table}
\begin{table}[!t]
\centering

\begin{tabular}{lll}
\hline
Diagnosis & Guesswork  & ReId AUC \\ 
\hline
\multicolumn{3}{c}{\name} \\
\hline
Edema         & $3617 $ $(94, 11544)  $  & $0.50$ $(0.49, 0.51)$\\
Consolidation & $1697 $ $(83, 5297)   $  & $0.55$ $(0.53, 0.57)$ \\
Cardiomegaly  & $9834 $ $(2072, 15766)$  & $0.51$ $(0.49, 0.53)$ \\
Atelectasis   & $13189$ $(2511, 28171)$  & $0.50$ $(0.48, 0.52)$ \\
\hline
\multicolumn{3}{c}{\textbf{Ablation}: \name with no label encoding ($T^X$ only)} \\
\hline
Edema         & $47$ $(12, 83)$ & $0.76$ $(0.76, 0.76)$ \\
Consolidation & $36$ $(2, 104)$ & $0.76$ $(0.76, 0.76)$\\
Cardiomegaly  & $42$ $(17, 57)$ & $0.75$ $(0.75, 0.75)$ \\
Atelectasis   & $80$ $(65, 98)$ & $0.75$ $(0.75, 0.75)$ \\

\hline
\end{tabular}
\caption{Privacy evaluation of \name when released with different diagnoses in MIMIC-CXR dataset. 
Metrics are averages over 100 trials using 10,000 samples each, followed by 95\% CI. 
}%
\label{tab:real-privacy}
\end{table}
\begin{table}[!t]
\centering

\begin{tabular}{lcccccc}
\hline
\textit{Encoding}  & E & Co & Ca & A & \textit{Avg}\\
\hline
Using raw data         & $0.91$ & $0.78$ & $0.89$ & $0.85$ & $0.86$\\
\hline
Using encoded data \\
\quad DP-S, $b=10$ & $0.51$ & $0.51$ & $0.52$ & $0.52$ & $0.52$\\
\quad DP-S, $b=20$ & $0.50$ & $0.50$ & $0.50$ & $0.50$ & $0.50$\\
\quad DP-S, $b=30$ & $0.49$ & $0.49$ & $0.50$ & $0.51$ & $0.50$\\
\quad DP-I, $b=1$   & $0.60$ & $0.59$ & $0.60$ & $0.59$ & $0.60$\\
\quad DP-I, $b=2$   & $0.54$ & $0.50$ & $0.55$ & $0.55$ & $0.54$\\
\quad DP-I, $b=5$   & $0.53$ & $0.55$ & $0.51$ & $0.52$ & $0.53$\\
\quad \name-Random      & $0.89$ & $0.75$ & $0.86$ & $0.84$ & $0.84$\\
\quad \name             & $0.82$ & $0.69$ & $0.81$ & $0.78$ & $0.78$\\
\hline
\end{tabular}
\caption{Utility for chest X-ray prediction tasks across different encoding schemes. All metrics are ROC AUCs across the MIMIC-CXR test set.  Guides of abbreviations for medical diagnosis: (E)dema, (Co)nsolidation, (Ca)rdiomegaly and (A)telectasis.}
\label{tab:utility-mimic}
\end{table}

\paragraph{Generalized Privacy} We first evaluated the guesswork and re-identification AUC (ReID AUC) of attackers trained using only encoded images (i.e. without labels) on the entire unfiltered MIMIC-CXR training set. For \name, this only requires using the neural encoder $T^X$.
We compared \name to prior lightweight encoding schemes, including InstaHide ~\cite{instahide} and Dauntless~\cite{DAUnTLeSS,xiao2021art}; and differential privacy methods, like DP-Image \cite{liu2021dp}. We now detail our baseline implementations.

\begin{itemize} %
    \item To assess the value of training \name's obfuscator layers, we  compared \name to an ablation with randomly initialized obfuscator layers, \name-Random.
    \item InstaHide randomly mixes each private image with 2 other private images (i.e with MixUp~\cite{zhang2018mixup}) and then randomly flips each pixel sign.
    \item Dauntless~\cite{xiao2021art} applies a separate random linear layer to each 16x16 pixel patch of the images, with each random weight initialized as according to a standard Gaussian distribution.
    \item DP-Simple adds independent laplacian noise to each pixel of the image to obtain differential privacy. We evaluated using a scale (or diversity parameter) $b$ of $10.0$, $20.0$ and $30.0$.
    \item DP-Image~\cite{liu2021dp} adds independent laplacian noise to the latent space of an auto-encoder to produce differentially private images.
      Our auto-encoder architecture is further detailed in Appendix \ref{sec:dp_image}.
      We trained our auto-encoder on the NIH dataset and applied it with laplacian noise on the MIMIC-CXR dataset. We evaluated using a scale $b$ of $1.0$, $2.0$ and $5.0$.
\end{itemize}

We report the expected guesswork and AUC for each attack as well as 95\% confidence intervals (CI). To compute confidence intervals, we sampled 100 bootstrap samples of 10,000 images (all encoded by a single $T$) from the MIMIC-CXR training set. Our 100 bootstraps consisted of 10 random data samples (of 10,000) across 10 random $T$.

\paragraph{Privacy with Real Labeling Functions}  In practice, the encoded images are released with encoded labels to enable model development on tasks of interest. Using this additional knowledge, attackers may be able to better re-identify private data. To evaluate the privacy of \name encodings when released with public labels, we trained the attackers to re-identify raw images given access to (raw image, raw label) pairs and (obfuscated image, obfuscated label) pairs.
To highlight the importance of \name's label encoding scheme $T^Y$ in this scenario, we also train attackers on an ablation of \name which does not encode the labels and releases (obfuscated image, raw label). This corresponds to using only \name's neural encoder $T^X$.

We performed this attack independently per diagnosis. We implemented the instance encoder $E^{\text{ins}}$ of our attacker as an SAU, our self-attention module, and represented the disease label an additional learned 256 dimensional input token for $E^{\text{ins}}$. As before, our attackers were trained for 500 epochs, and evaluated on the MIMIC-CXR training set. We report the expected guesswork and AUC for each attack as well as 95\% confidence intervals. To compute confidence intervals, we sampled 100 random $T$ and encoded the whole class-balanced MIMIC-CXR training set for each sampled $T$.

\paragraph{Utility Evaluation}
We evaluated the utility of an encoding scheme on the MIMIC-CXR dataset by measuring the ROC-AUC of diagnosis models trained using its encodings. We compared the utility of \name to a plaintext baseline (i.e. using raw data), which provides us with a utility upper bound. To isolate the impact of training \name's obfuscator layers on utility, we also compared the utility of \name to \name-Random. We also computed the utility of our differential privacy baselines, DP-Simple with a scale parameter $b$ of 10, 20 and 30 and DP-Image with a scale of 1, 2 and 5. For each encoding scheme, we experimented with different classifier architectures (e.g. SAU vs ResNet-18), dropout rates and weight decay, and selected the architecture that achieved the best validation AUC.

\section{Results}

\paragraph{Generalized Privacy} We report our generalized privacy results, which consider re-identification attacks on the unlabeled MIMIC-CXR dataset, in Table \ref{tab:gen-privacy} and Table \ref{tab:cross-attacker-privacy}, with higher guesswork and lower ReID AUC denoting increased privacy.
While \name was trained to maintain privacy against an SAU-based attacker on the NIH training set, we found that its privacy generalized to a held-out dataset, MIMIC-CXR, and held-out attack architectures (e.g. ResNet-18 and ViT). \name obtained a guesswork of 8411 (95\% CI 5219, 12033) and an ReId AUC of 0.50 (95\% CI 0.49, 0.51) against a ViT attacker. We note that a guesswork of 10,000 corresponds to guessing randomly in this evaluation. In contrast, the InstaHide and Dauntless baselines could not defend against re-identification attacks obtaining both a guesswork of 1.0 (95\% CI 1, 1).
As illustrated in Appendix~\ref{sec:viz_encodings},
the differential privacy baselines can obtain privacy at the cost of significant image distortion. DP-Image with a laplacian noise scale of 5.0 obtained a guesswork of 1379 (95\% CI 49, 4135) and an attacker AUC of 0.59 (95\% CI 0.59, 0.60).

\paragraph{Privacy with Real Labeling Functions} We evaluated the privacy of releasing \name encodings with different public labels in Table  \ref{tab:real-privacy}.
Releasing \emph{raw labels} resulted in significant privacy leakage with guessworks ranging from 36 (95\% CI 2, 104) to 80 (95\% CI 65, 98) for Consolidation and Atelectasis respectively.
In contrast, when labels are protected using \name's label encoding scheme and released alongside the image encodings, \name maintains privacy across all diagnoses tasks, with guessworks ranging from 1697 (95\% CI 83, 5297) to 13189 (95\% CI 2511, 28171) for Consolidation and Atelectasis respectively.

\paragraph{Utility Evaluation}
We report our results in predicting various medical diagnoses from X-rays in Table \ref{tab:utility-mimic}. Models built on \name obtained an average AUC of 0.78, compared to 0.86 by the plaintext baseline and 0.84 by the \name-Random baseline. In contrast, the best differential privacy baseline, Image-DP, obtained average AUCs of 0.60, 0.54 and 0.53 when using a scale of $1$ and $2$ and $5$ respectively. \name obtained a 25 point average AUC improvement over DP-Image while obtaining better privacy.

\section{Conclusion}
We propose \name, an encoding scheme for releasing private data for machine learning model development while preventing raw data re-identification. 
\name uses trained obfuscator layers and random neural networks to minimize the likelihood of re-identification, while encouraging the invertability of the overall transformation. 
In experiments on MIMIC-CXR, a large chest X-ray benchmark, we show that \name obtains strong privacy across held-out attackers, obtaining an average guesswork of 8411, whereas prior encoding schemes like Dauntless~\cite{DAUnTLeSS}, InstaHide~\cite{instahide} did not meet our privacy standard, obtaining guessworks of 1.  
While differential privacy baselines can achieve privacy with enough noise, we found this came with a massive loss of utility, with DP-Image obtaining an average AUC of 0.53 for a guesswork of 1379. In contrast, models built on \name encodings approached the utility of our plaintext baseline, obtaining an average AUC of 0.78 compared to 0.86 by the plaintext model. %

\textbf{Future Work} %
While our threat model considers a computationally unbounded adversary, in practice, we rely on model-based attackers for both the development and evaluation of \name. More powerful models may result in more successful attacks on \name. As a result, continued research into re-identification algorithms is needed to offer stronger theoretical guarantees and develop more powerful encodings. 
Moreover, while we show that \name generalizes to an unseen datasets, %
this does not guarantee that it will generalize to arbitrary datasets. Additional research studying the privacy impact of domain shifts is also necessary.

\bibliography{references.bib}
\bibliographystyle{icml2021}

\clearpage

\appendix

Our code will be publicly released after the review process. 

\section{Guesswork Supplementary Details}
\label{sec:guesswork}

Recall that for an ordered list of $mn$ correspondence guesses $(u_1, \dots, u_{mn})$, where $u_i \in X_E \times Z_A$, the guesswork is defined as the rank of the first correct guess:  $ \mathcal{G} = \min_k \{k \text{ s.t. } u_k \in M_T\}$, where  $M_T = \{ (x, z) \in X_E \times Z_A   \text{ s.t. } z = T^X(x) \}$.  In the event of ties, the guesswork is computed as the expected value over permutations of the suitable subsets. In the paper, we use $X_E=X_A$ but the guesswork can be computed for an arbitrary superset $X_E \supseteq X_A$ of size $m$.

\paragraph{Guesswork Algorithm}

We propose the following algorithm to compute the guesswork for a given probability matrix and set of correct guesses.

\begin{algorithm}[H]
\caption{Guesswork algorithm}
\label{alg:guesswork}
\fontsize{9.5}{11.4}\selectfont
\hspace*{1pt} \textbf{Input} Correct matching $M_T = \{(x_i,z_j) \text{ s.t. } z_j = T^X(x_i)$\}  \\
\hspace*{1pt} \textbf{Input} Probability matrix $A$ where $A_{i,j} = \mathbb{P}((x_i, z_j) \in M)$ \\
\hspace*{1pt} \textbf{Output} Guesswork $\mathcal{G}$ for $A$
\begin{algorithmic}[1]
 \STATE From $A$, extract \\
 $S =\{(i, j, A_{i,j}) \text{ for } 1\le i \le m, 1\le j \le n\}$
 \STATE Partition $S$: \\
 $S = \bigcup_p S_p$ where $S_p = \{(i, j, A_{i,j}) \text{ s.t. } A_{i,j}=p\}$
 \STATE Find the highest value of $p$ such that $A_p$ contains matches: \\
 $q = \max_p \{p \text{ s.t. } \exists (i,j, A_{i,j}) \in S_p \text{ s.t. }  (x_i, z_j) \in M_T\}$
 \STATE $\mathcal{G} \gets 0$
 \FOR{$p > q$}
    \STATE  $\mathcal{G} \gets \mathcal{G} + |A_p|$
 \ENDFOR
\STATE  $\mathcal{G} \gets \mathcal{G} + \frac{1 + |A_q|}{1 + |A_q \cap M| }$
\STATE \textbf{return} $\mathcal{G}$

\end{algorithmic}
\end{algorithm}

The expression $\frac{1 + |A_q|}{1 + |A_q \cap M|}$ is derived by computing the expected value of number of trials before success in the urn problem without replacement.

\begin{example}[guesswork calculation extended]
    Let $X_E = X_A = \{1, 2\}$ and disregard labels for now. Consider three transformations $X_A \rightarrow \{a,b,c,d\}$:
    
    $
   T_1: \left\{
        \begin{array}{ll}
            1 & \mapsto a \\
            2 & \mapsto b
        \end{array}
    \right.
    \quad
    T_2: \left\{
        \begin{array}{ll}
            1 & \mapsto b \\
            2 & \mapsto a
        \end{array}
    \right.
    \quad
    T_3: \left\{
        \begin{array}{ll}
            1 & \mapsto c \\
            2 & \mapsto d
        \end{array}
    \right.
    $
    
    We evaluate the following encoding schemes, defined by the 
    distribution used to sample $T$:
    
    $\Gamma_1:\quad 
    \mathbb{P}(T_1)  =2/3, 
    \mkern9mu \mathbb{P}(T_2) = 1/3, \mkern9mu \mathbb{P}(T_3) = 0 $
    
    $\Gamma_2:\quad 
    \mathbb{P}(T_1) = 1/2, 
    \mkern9mu \mathbb{P}(T_2) = 1/2, \mkern9mu \mathbb{P}(T_3) = 0 $
    
    $\Gamma_3:\quad 
    \mathbb{P}(T_1)  =1/3, 
    \mkern9mu \mathbb{P}(T_2) = 1/3, \mkern9mu \mathbb{P}(T_3) = 1/3 .$
    
 For $\Gamma_1$, Eve observes $Z_A = \{a, b\}$ regardless of the choice of $T_A$. Given her knowledge of $\mathbb{P}(\bm{T})$, she elects to rank  $\{(1, a), (2,b)\}$ before $\{(1,b), (2,a))\}$ which gives guessworks $\mathcal{G}(T_1) = 1$ and $\mathcal{G}(T_2) = 3$. In expectation, the guesswork of $\Gamma_1$ is 5/3 (with a variance of 8/27). 
 
 For $\Gamma_2$, Eve observes $Z_A = \{a, b\}$ as well, but equally ranks all $4!$ orderings of the guesses $((1, a), (1,b), (2,a), (2,b))$, which leads to the same guesswork for both $T$: $\mathcal{G}(T_1) = \mathcal{G}(T_2) =  \frac{1}{2}\cdot 1 + \frac{1}{2}\cdot\frac{2}{3}\cdot 2 + \frac{1}{2}\cdot\frac{1}{3}\cdot 3 = \frac{5}{3}$ (no variance). 
 
 For $\Gamma_3$, whenever Eve observes $Z_A = \{c, d\}$, she deduces that $T_A=T_3$, which leads to a guesswork of 1. In the other cases, observing $Z_A = \{a, b\}$ means that $T_1$ and $T_2$ are equally likely, so the guesswork is 5/3. In expectation, the guesswork is 13/9, which is lower (and thus worse privacy) than the previous schemes.

\end{example}

\paragraph{Guessworks in Special Cases} We discuss two special cases that arise when computing guesswork.
\begin{enumerate}
    \item If all guesses in the bucket $A_p$ of highest probability are correct guesses, then the guesswork is 1, characterizing a non-private scheme. Note that this does not depend on the cardinal of the bucket $A_p$ of highest probability: regardless of whether the attacker is confidently correct about one matching pair or multiple matching pair, the guesswork will still be 1.
    \item If the probability matrix is uniform (i.e. there is $p$ for which $S = S_p$, such that all guesses are in the same bucket), then the guesswork is $\frac{mn+1}{n+1}$, i.e. $\mathcal{G} \approx |X_E|$. This characterizes an attacker that fails to capture any privacy leakage of the encoding scheme.
\end{enumerate}

Note that $|X_E|$ is not an upper-bound of guesswork. An attacker that is confidently wrong can achieve a guesswork up to $mn-n+1$.

\paragraph{Discussion on Eve's strategy} In our definition of guesswork, Eve commits to a probability matrix $A_{i,j} = \mathbb{P}((x_i, z_j) \in M)$, then enumerates her guesses in descending order of likeliness. This would not be the optimal strategy for an attacker who wishes to minimize the number of guesses required to identify a correct match. For instance, if $\mathbb{P}((x_i, z_j)$ is uniform, Eve could commit to a single column (or row) and achieve an expected number of guesses of $m/2$ (or $n/2$). More generally, after Eve made her first guess $u_1$, she can assume the first guess was incorrect and recompute the new probability matrix $\mathbb{P}((x_i, z_j) \in M | u_1 \not \in M)$, then proceed with subsequent guesses. Such an auto-regressive strategy is costly to implement. In practice, Eve also would not have access to an oracle that notifies her when a guess is correct. Therefore, we adopt the definition of guesswork exposed in Section~\ref{sec:problem_statement} as an efficient way to universally compare the privacy of different encoding schemes.

\section{Dataset Statistics}
\label{sec:dataset_stats}

We leveraged the NIH training set for training \name, and leveraged the unlabeled MIMIC-CXR training set for all generalized privacy evaluation. To evaluate utility and privacy with real labeling functions, we use the labeled subsets of the MIMIC-CXR dataset. The labeled MIMIC-CXR training and validation sets were filtered to be class balanced, by assigning random one negative control for each positive sample. The number of images per dataset is shown in Table \ref{tab:data}

\begin{table}[H]
\centering
\label{tab:data}
\begin{tabular}{cccc}
\hline
Dataset  & Train & Dev & Test\\ 
\hline
\multicolumn{4}{c}{\textit{Unlabeled}} \\
\hline
NIH &  40365 & NA & NA  \\
MIMIC-CXR & 57696 & NA & NA  \\
\hline
\multicolumn{4}{c}{\textit{Labeled}} \\
\hline
MIMIC-CXR E &  3660 & 1182 & 12125 \\
MIMIC-CXR Co & 1120 & 375 & 11031  \\
MIMIC-CXR Ca & 11724 & 3876 & 12791  \\
MIMIC-CXR A &   2164 & 3992 & 12129  \\

\hline
\end{tabular}
\caption{Dataset statistics for all datasets. The training and development sets of MIMIC CXR Edema, Consolidation, Cardiomegaly and Atelecatasis were filtered to contain one negative control for each positive sample. Guides of abbreviations for medical diagnosis: (E)dema, (Co)nsolidation, (Ca)rdiomegaly and (A)telectasis.}
\end{table}

\section{SAU: Simple Attention Unit}
\label{sec:sau}
\begin{figure}[H]
    \centering
    \includegraphics[width=0.5\textwidth]{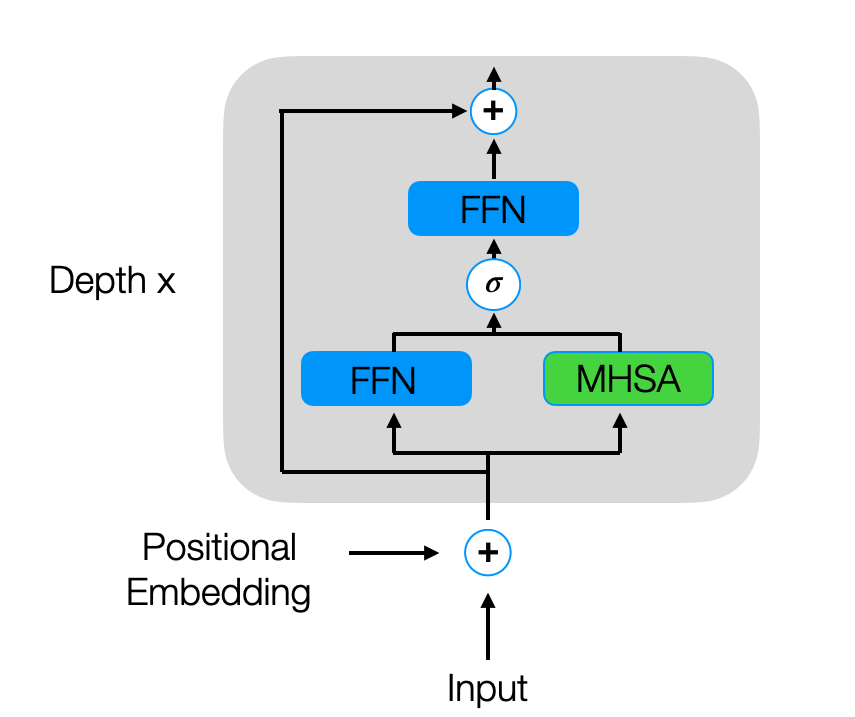}
    \caption{Simple Attention Unit Architecture. The module uses a learnable gate at each layer to interpolate between leveraging behaving as a feed forward network (FFN) and a multi-headed self attention network (MHSA).}
    \label{fig:sau}
\end{figure}

Our Simple Attention Unit (SAU), illustrated in Figure \ref{fig:sau}, utilizes a learned gate, $\alpha$, at each layer to interpolate between acting as a standard feed forward network (FFN) with no attention computation, and a multi-head self-attention (MHSA) network. We found that this allowed for faster and more stable training compared to ViTs\cite{zhou2021deepvit, dosovitskiy2020image} in both privacy and utility experiments.  To encode patch positions, we leverage a learned positional embedding for each location, following prior work \cite{zhou2021deepvit, dosovitskiy2020image}. Each layer of the SAU is composed of the following operations:

$$ x_{norm} = \text{BatchNorm}(x)$$
$$ h_{ffn} = \text{SELU}( W_{in} x_{norm}  + b_{in}  )$$
$$ h_{attn} =  \text{MHSA}(x_{norm})$$
$$ h = \sigma(\alpha) \times h_{attn} + (1- \sigma(\alpha)) \times h_{ffn}$$
$$ o =  \text{SELU}( W_{o}h + b_{o}) + x_{norm}   $$

Where Multi-head self-attention ($\text{MHSA}$) is defined as:
$$K_i, Q_i, V_i= W_{i} x_{norm}$$
$$ \text{head}_i =  softmax(\frac{Q_i  K_i^T}{\sqrt{d_k}}) V^i $$
$$ h_{attn} = \text{BatchNorm}( Concat(head_1, ..., head_h)W_{attn}  )$$

Where $d_k$ is the dimension of each head, all $W$ and $b$ are learned parameters, and $\alpha$ is a learned gate. $\alpha$ is initialized at $-2$ for each layer.

\section{DP-Image Baseline}
\label{sec:dp_image}

DP-Image\cite{liu2021dp} is a differential privacy method based adding laplacian noise to the latent space of auto encoders to achieve differential privacy.
We trained our auto-encoder on the NIH training set, with no noise, and apply it with noise on the MIMIC dataset. Our encoder, mapping each 64x64 pixel image $x$ to a 256d latent code  $z$, is composed of six convolutional layers, each followed by a leaky relu activation and batch normalization. Each convolutional layer had a kernel size of 3, a stride of 2. This was then reduced a single code $z$ with global average pooling. Our decoder, which mapped $z$ back to $x$, consisted of six transposed convolutional layers, each followed by a leaky relu activation and batch normalization. The auto encoder was trained to minimize the mean squared error between the decoded image and the original image.

\section{Visualizations}
In Figure \ref{fig:distortion}, we visualize the impact of \name and DP-Image encodings when using different amounts of noise (parametrized by the diversity parameter $b$). Each row represents a different image. The \textit{raw\_x} column are raw images. The \name column shows encodings obtained when applying \name's neural encoder for a specific choice of private weights $\theta_{key}$: those are representative of the released images. We then train a decoder $D_T$ for a specific choice of private weights $\theta_{key}$. During training, the decoder has access to parallel data (raw image, encoded \name image). We visualize the decoded images in the \textit{\name decoded} column. Note that in our scenario, only Alice would be able to train such a decoder: Bob and Eve only have access to encoded images with labels. The \textit{DP-image no noise} column is the reconstructed image obtained with the trained auto-encoder that is used for the DP-Image baseline. We also visualize the reconstructed images when varying amounts of noise are added.
\onecolumn
\begin{landscape}
\label{sec:viz_encodings}
\begin{figure}[H]
    \centering
    \includegraphics[height=0.9\textwidth]{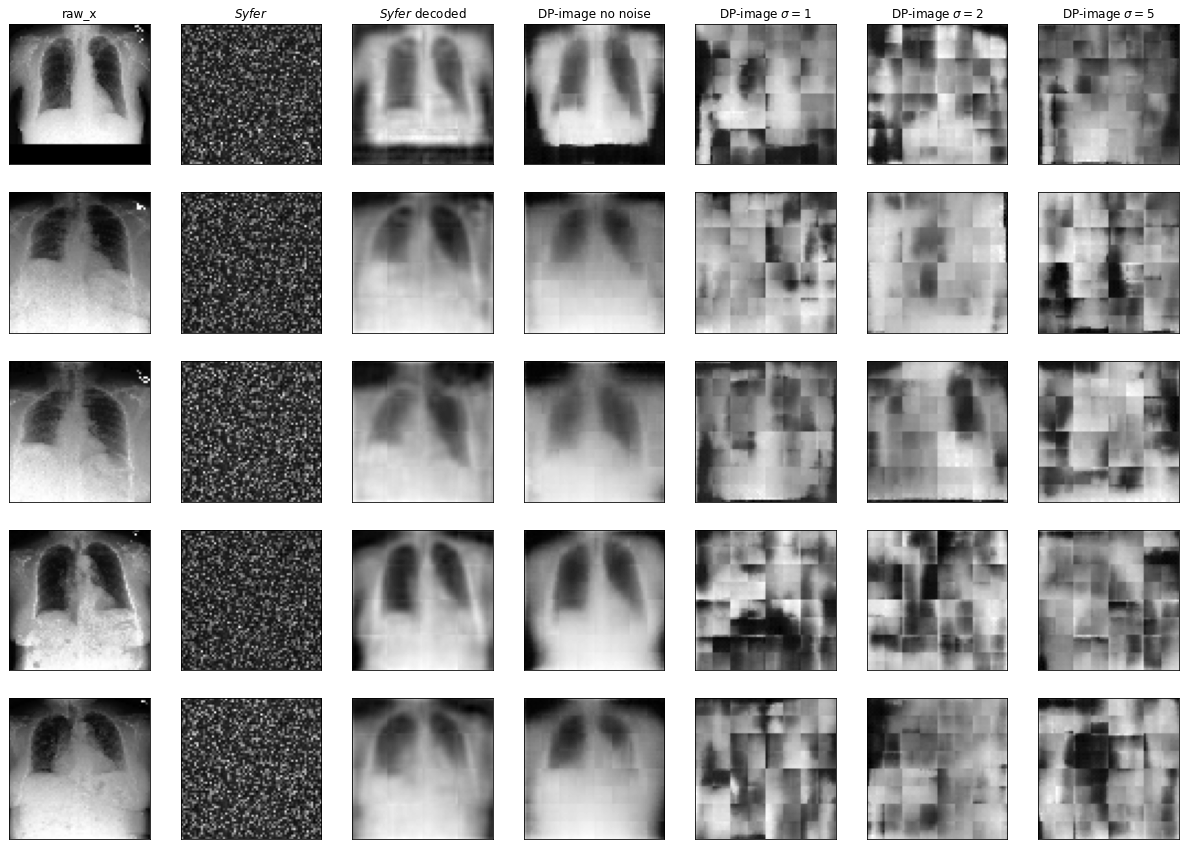}
    \caption{Vizualisations of raw images, \name encodings, decoded \name encodings, and DP-Image encodings. \name encodings were obtained after applying the $T^X$ part of \name to raw images. Decoded \name encodings are obtained by a model $D_T$ trained on a set of parallel training data (plaintext attack). DP-Image encodings are shown with varying amount of noise.}
    \label{fig:distortion}
\end{figure}

\end{landscape}
\newpage
\twocolumn
\section{Additional Privacy Analyses}
\label{sec:addit_privacy}

\begin{table}[!htb]
\centering
\begin{tabular}{lcc}
\hline
Patch size & Guesswork & ReId AUC\\ 
\hline
\multicolumn{3}{c}{\name} \\
\hline
32px & 102795 (25221, 235114) &  50 (50, 50)\\
16px & 12715 (4670, 31748) & 50 (46, 54) \\
\hline
\end{tabular}
\caption{Privacy evaluation of \name against attackers attempting to re-identify patches of the encoded images. Guesswork was computed over a random subset of 10,000 samples. All metrics are followed by 95\% confidence intervals.}
\label{tab:patch-privacy}
\end{table}

\section{Additional Utility Analyses}
\label{sec:addit_utility}

\begin{figure}[H]
    \centering
    \includegraphics[width=0.5\textwidth]{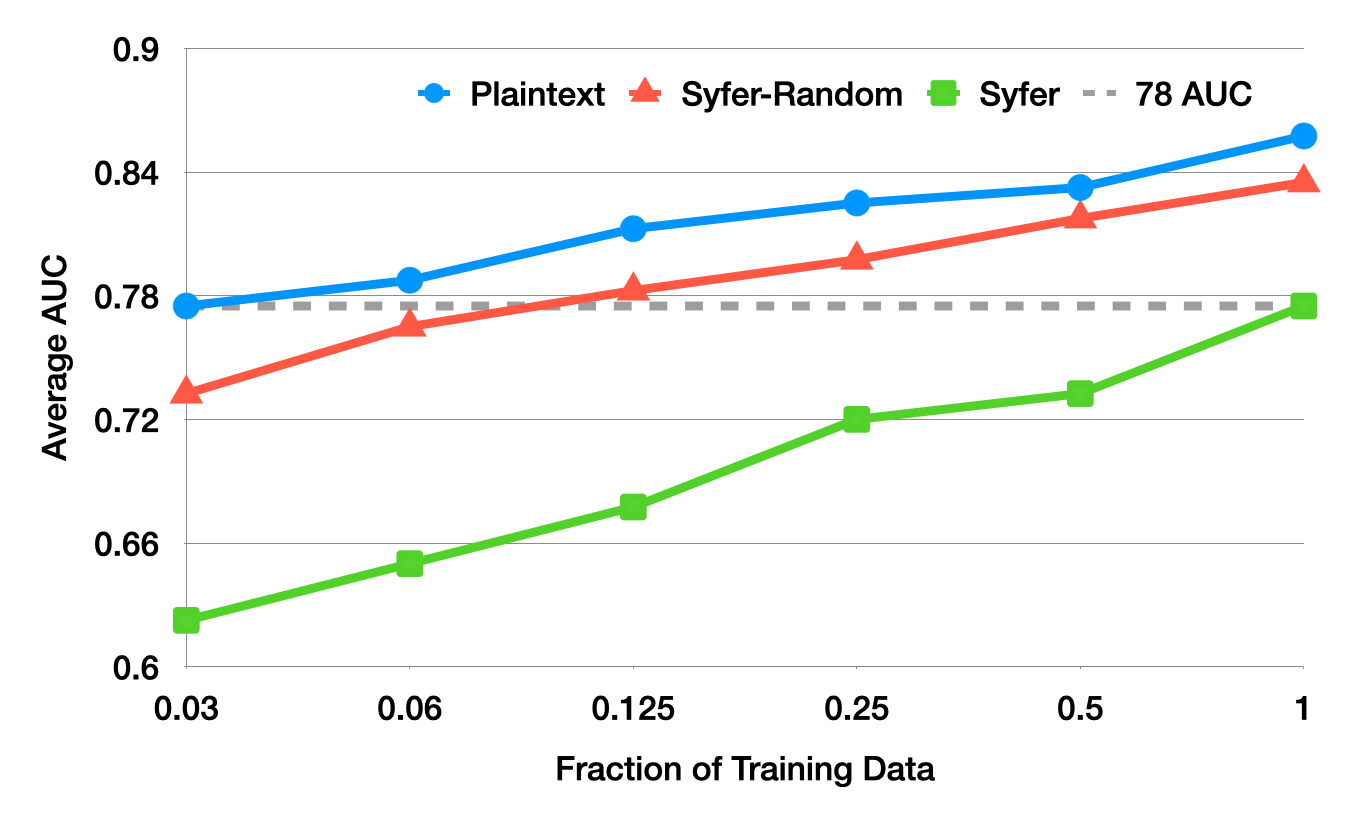}
    \caption{Average AUC on MIMIC-Test set when training with different fractions of the data when using \name, \name-Random, and Plaintext encodings. }
    \label{fig:learning_curve}
\end{figure}

In Figure \ref{fig:learning_curve}, we plot the learning curves of \name, \name-Random and our plaintext baselines when training on fractions $\frac{1}{32}$, $\frac{1}{16}$, $\frac{1}{8}$, $\frac{1}{4}$, $\frac{1}{2}$ and $1$ of the data. 

We find that it takes plaintext models $\frac{1}{2}$ of the training data to reach the full performance of \name-Random, indicating that using a random \name architecture harms sample complexity. \name, which achieves strong privacy, requires more data to achieve the same utility, with \name-Random achieving the same average AUC when using less than $\frac{1}{8}$ of the data and Plaintext achieving the same performance when using $\frac{1}{32}$.

\end{document}